\documentclass[letterpaper, 10 pt, conference]{ieeeconf}  
\UseRawInputEncoding

\IEEEoverridecommandlockouts                              
\usepackage{subcaption}
\usepackage{cite}
\usepackage{amsmath,amssymb,amsfonts}
\usepackage{algorithm}
\usepackage{algpseudocode}
\usepackage{graphicx}
\usepackage{textcomp}
\usepackage{xcolor}
\usepackage{url}
\usepackage{bm}
\usepackage[table]{xcolor}
\usepackage{booktabs}
\usepackage{multirow}
\usepackage[hidelinks]{hyperref}

\title{\LARGE \bf
	TopoMix: Continuous Spatial Mixing for Long-Tailed Data Augmentation
}

\author{Yefan Wang$^{1,*}$, Yusen Wu$^{2}$, Zihui Ling$^{3}$, and Jianyuan Ni$^{4}$
	\thanks{$^{1}$Yefan Wang is with the University of Shanghai for Science and Technology, 
		Shanghai, 200093, China,
		{\tt\small yefanwang88@gmail.com}}%
	\thanks{$^{2}$Yusen Wu is with the Fujian University of Technology, 
		Fujian, 350118, China, 
		{\tt\small 3231319130@smail.fjut.edu.cn}}%
	\thanks{$^{3}$Zihui Ling is with the Shanghai DianJi University, 
		Shanghai, 201306, China, 
		{\tt\small a1711934954@gmail.com}}%
	\thanks{$^{4}$Jianyuan Ni is with the Juniata College, 
		Huntingdon, PA 16652, USA,
		{\tt\small jni100@juniata.edu}}
	\thanks{$^{*}$Corresponding author: Yefan Wang.}
}

\begin{document}

\maketitle
\thispagestyle{empty}
\pagestyle{empty}

\begin{abstract}
	Most mixing-based augmentation methods improve generalization through a simple idea: combine two training samples with a fixed global ratio or a hard spatial replacement. While effective, this design leaves little room for local variation, which becomes a more visible limitation in long-tailed recognition, where minority classes are especially sensitive to the quality of augmented samples. We propose TopoMix, a training-free augmentation method that replaces fixed mixing with continuous spatial mixing. TopoMix constructs a smooth spatially varying mask from a low-resolution random control grid, bicubic interpolation, and stochastic non-linear shaping, and uses this mask for element-wise image mixing and label reweighting. Without introducing trainable parameters or auxiliary networks, TopoMix offers a simple way to make mixing more flexible in the input space. Experiments on standard long-tailed benchmarks show that it consistently improves over representative mixing-based baselines, indicating that smooth spatial mixing is an effective augmentation strategy for long-tailed recognition.
\end{abstract}

\section{Introduction}
\label{sec:intro}

Deep neural networks have achieved strong performance across a wide range of learning problems, including visual recognition, sequence modeling, multimodal sensing, healthcare analysis, and environmental assessment \cite{he2016deep,dosovitskiy2020image,deng2009imagenet,ding2025advances,qin2025kedformer,feng2025machine,ni2022cross,ni2022progressive,ni2024survey,ni2025s3}. Despite this progress, model quality still depends heavily on the scale, diversity, and balance of training data \cite{krizhevsky2009learning}. In practical settings, limited data, distribution imbalance, and weak minority-class supervision often lead to overfitting and poor generalization \cite{maray2023transfer,liu2019large,kang2019decoupling,alshammari2022long,8953804,ren2020balanced}. Data augmentation is a standard way to improve generalization under such conditions. Automated strategies such as AutoAugment \cite{cubuk2019autoaugment} and RandAugment \cite{cubuk2020randaugment} learn or simplify augmentation policies, while mixing-based methods provide a particularly simple and effective form of regularization. Mixup \cite{zhang2017mixup} linearly interpolates two samples with a single global ratio, and CutMix \cite{yun2019cutmix} replaces a hard image region with a patch from another sample. For imbalanced learning, Remix \cite{chou2020remix} and Balanced-Mixup \cite{galdran2021balanced} further modify the mixing process to better account for minority classes, while Manifold Mixup \cite{verma2019manifold} performs interpolation in hidden space instead of the input space. More recent methods, such as Text-Guided Mixup \cite{DBLP:conf/bmvc/FranklinYZ0H24} and ConCutMix \cite{10599153}, further introduce semantic or contrastive guidance for long-tailed mixing.

\begin{figure*}[t]
	\centering
	\vspace*{5pt}
	\includegraphics[width=0.99\textwidth]{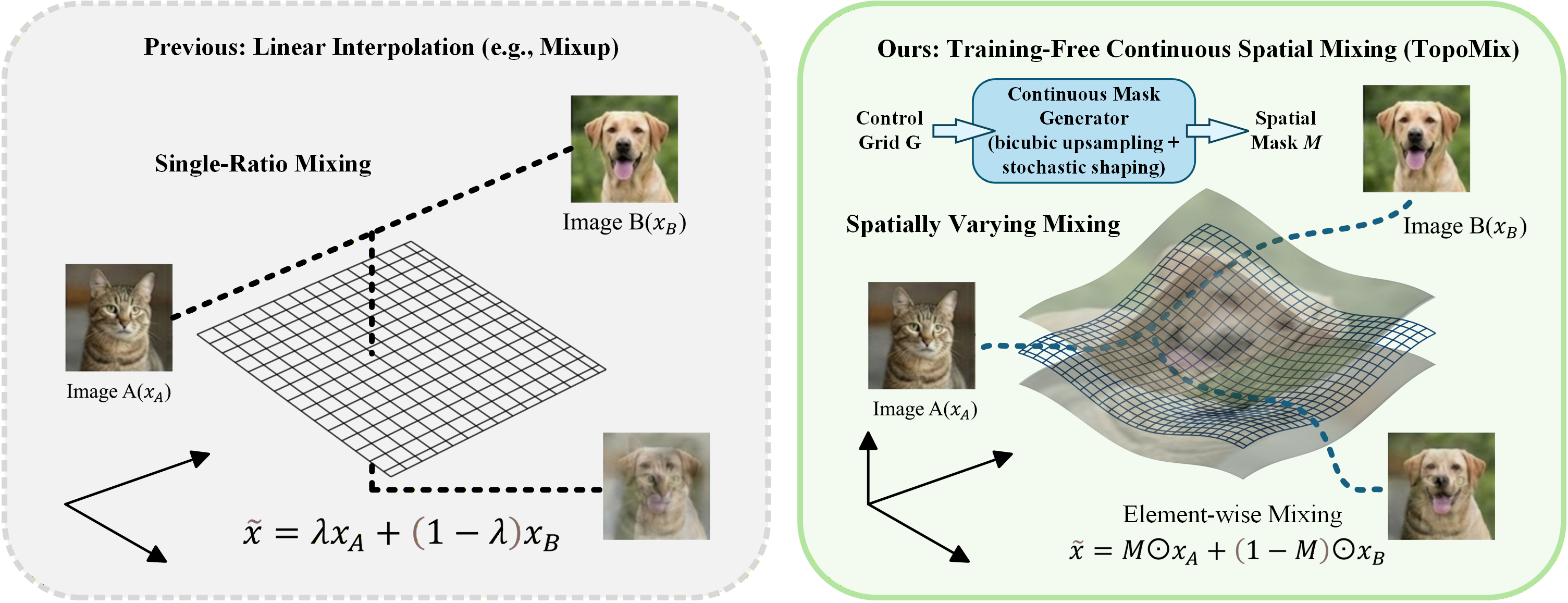}
	\caption{Left: Global linear mixing. Right: TopoMix with smooth spatial mixing.}
	\label{fig:framework}
\end{figure*}

However, most existing mixing methods still rely on either a single global interpolation ratio or hard spatial replacement. This design is restrictive for long-tailed recognition. Minority classes are more sensitive to unstable decision boundaries and weak feature formation, so unrealistic or overly rigid mixed samples may provide limited benefit. From the perspective of manifold structure and intrinsic dimensionality, natural images are not well described by a single linear path between two samples \cite{fefferman2016testing,narayanan2010sample,fukunaga1971algorithm,gong2019intrinsic,pope2021intrinsic}. This motivates a simple question: can we keep mixing while making it spatially continuous and locally varying? To answer this question, we propose \textbf{TopoMix}, a training-free augmentation method based on continuous spatial mixing. TopoMix first samples a low-resolution random control grid, upsamples it by bicubic interpolation, and then applies a stochastic non-linear shaping function to obtain a smooth spatially varying mask. The mask is used for element-wise image mixing, and the target label is reweighted by the mask average. In this way, TopoMix replaces fixed global mixing with continuous local variation, without introducing trainable parameters or auxiliary networks. An overview of the method is shown in Fig.~\ref{fig:framework}. Our contributions are summarized as follows: (1) We propose TopoMix, a training-free data augmentation method for long-tailed recognition based on continuous spatial mixing. (2) We design a simple mask generation pipeline based on a random control grid, bicubic interpolation, and stochastic non-linear shaping, which produces smooth spatially varying masks for element-wise mixing. (3) We evaluate TopoMix against diverse representative mixing-based augmentation methods on standard long-tailed benchmarks. The results show that TopoMix consistently achieves the best overall performance, and further validation on both ResNet-32 and ResNet-50 demonstrates that the proposed strategy remains effective across different backbone settings.

\section{Related Work}
\label{sec:related}

\subsection{Mixing-based Data Augmentation}
Data augmentation is a standard tool for improving generalization. Early methods mainly relied on handcrafted transformations, and later works such as AutoAugment \cite{cubuk2019autoaugment} and RandAugment \cite{cubuk2020randaugment} further simplified or automated policy design. Among them, mixing-based augmentation has become particularly influential due to its simplicity and strong regularization effect. Mixup \cite{zhang2017mixup} generates virtual samples by linearly interpolating two inputs and their labels with a single global ratio. CutMix \cite{yun2019cutmix} instead replaces a hard image region with a patch from another sample, which preserves local image content but introduces abrupt spatial transitions. Manifold Mixup \cite{verma2019manifold} moves interpolation from the input space to hidden representations to improve feature learning. For imbalanced settings, Remix \cite{chou2020remix} adjusts the mixing labels according to class frequencies, and Balanced-Mixup \cite{galdran2021balanced} combines standard sampling with class-balanced sampling to better support minority classes. More recent methods such as Text-Guided Mixup \cite{DBLP:conf/bmvc/FranklinYZ0H24} and ConCutMix \cite{10599153} further introduce semantic or contrastive guidance for long-tailed mixing. Most existing mixing methods still rely on either a single global interpolation ratio or hard regional replacement. In contrast, TopoMix performs continuous spatial mixing in the input space through a smooth spatially varying mask, providing a simple alternative to fixed global mixing while introducing richer local variation.

\subsection{Long-Tailed Recognition and Manifold Perspective}
Long-tailed recognition remains challenging because head classes dominate optimization, while tail classes often suffer from weak representations and poor decision boundaries \cite{liu2019large,kang2019decoupling,cao2019learning,alshammari2022long,8953804,ren2020balanced}. Beyond reweighting and classifier adjustment, recent long-tailed recognition studies have also explored bilateral-branch learning, logit adjustment, label-distribution disentangling, diverse expert routing, and calibrated expert ensembles \cite{Zhou_2020_CVPR,DBLP:conf/iclr/MenonJRJVK21,Hong_2021_CVPR,DBLP:conf/iclr/WangLM0Y21,Aimar_2023_CVPR}. Existing solutions mainly improve reweighting, margin design, classifier adjustment, or resampling, but the quality of augmented samples also plays an important role in how minority classes are learned.

From a geometric perspective, natural images are often assumed to concentrate around lower-dimensional structures rather than along simple linear paths \cite{fefferman2016testing,narayanan2010sample,fukunaga1971algorithm,gong2019intrinsic,pope2021intrinsic}. This view helps explain why globally linear mixing may be restrictive, especially in long-tailed settings where minority-class samples are already scarce and fragile. Our method does not attempt to explicitly model the data manifold. Instead, it introduces a smooth spatially varying mixing mask in the input space, providing a training-free alternative to fixed global mixing for long-tailed augmentation.

\section{Methodology}
\label{sec:method}

\subsection{Preliminaries: Mixing Forms}
Let $\mathbf{x}_A, \mathbf{x}_B \in \mathbb{R}^{C \times H \times W}$ be two samples drawn from the training distribution $\mathcal{D}$, with corresponding labels $y_A$ and $y_B$. A general mixing operator produces a synthetic sample $\tilde{\mathbf{x}}=\mathcal{M}(\mathbf{x}_A,\mathbf{x}_B,\Lambda)$, where $\Lambda$ denotes the mixing variable.

\begin{figure}[t]
	\centering
	\vspace*{5pt}
	\includegraphics[width=0.99\linewidth]{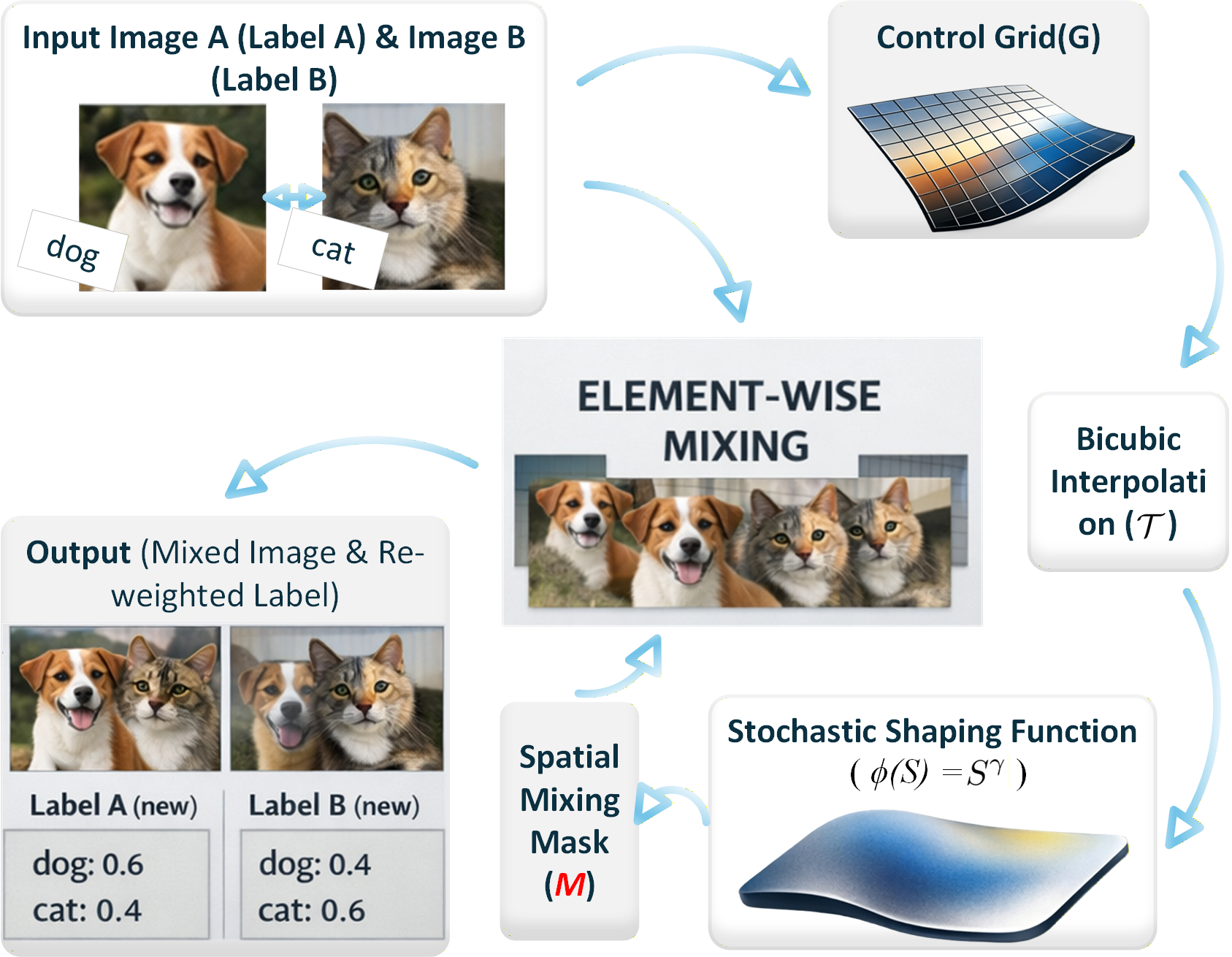}
	\caption{Overview of TopoMix. A random control grid is upsampled by bicubic interpolation and stochastically shaped to generate a spatial mixing mask for element-wise mixing.}
	\label{fig:frame}
\end{figure}

In Mixup, $\Lambda$ is a scalar $\lambda \in [0,1]$, and the mixed sample is obtained by global linear interpolation. This design is simple and effective, but it applies the same mixing ratio to all spatial locations. In CutMix, $\Lambda$ is a binary spatial mask, which preserves local image content but introduces abrupt region boundaries. These two forms represent two common input-space mixing patterns: a single global ratio and hard regional replacement. Our goal is to retain the simplicity of input-space augmentation while introducing smooth local variation into the mixing process. To this end, we construct a continuous spatial mask for element-wise mixing. An overview is shown in Fig.~\ref{fig:frame}.

\begin{algorithm}[htbp]
	\caption{TopoMix Augmentation}
	\label{alg:topomix}
	\begin{algorithmic}[1]
		\Require Input samples $\mathbf{x}_A,\mathbf{x}_B$, labels $y_A,y_B$, grid size $k$, probability $p$
		\Ensure Mixed sample $\tilde{\mathbf{x}}$ and label $\tilde{y}$
		
		\State Sample $u\sim\mathcal{U}(0,1)$
		\If{$u>p$}
		\State \Return $(\mathbf{x}_A,y_A)$
		\EndIf
		\State Sample a low-resolution control grid $\mathbf{G}\sim\mathcal{U}(0,1)^{k\times k}$
		\State Upsample $\mathbf{G}$ to image resolution by bicubic interpolation to obtain $S$
		\State Sample $\gamma\sim\mathcal{U}(0.5,2.0)$
		\State Generate the spatial mask $M=\mathrm{Clip}(S^\gamma,0,1)$
		\State Compute mixed sample $\tilde{\mathbf{x}} = M\odot\mathbf{x}_A + (1-M)\odot\mathbf{x}_B$
		\State Compute $\lambda_{\text{eff}}=\frac{1}{H\times W}\sum_{u,v}M_{u,v}$
		\State Compute mixed label $\tilde{y}=\lambda_{\text{eff}}y_A+(1-\lambda_{\text{eff}})y_B$
		\State \Return $(\tilde{\mathbf{x}},\tilde{y})$
	\end{algorithmic}
\end{algorithm}

\subsection{TopoMix: Continuous Spatial Mixing}
The name ``TopoMix'' comes from the topographic structure of the mixing mask. TopoMix constructs a spatially continuous mixing field over the image plane, where neighboring pixels have correlated mixing weights, as illustrated in Fig.~\ref{fig:frame}, making the mixed sample follow smooth local transitions rather than a single global ratio or discontinuous binary replacement. This topographic mixing field differs from the single global coefficient in Mixup and the discontinuous binary regions in CutMix, while preserving local spatial continuity during sample mixing.

\subsubsection{Random Control Grid and Bicubic Upsampling}
We first sample a low-resolution control grid $\mathbf{G} \sim \mathcal{U}(0,1)^{k \times k}$, where $k \ll H$ controls the spatial granularity of the mask. We then upsample $\mathbf{G}$ to the image resolution using bicubic interpolation: $S=\mathcal{T}_{\text{bicubic}}(\mathbf{G}, \text{size}=(H,W))$. This produces a smooth spatial surface $S \in [0,1]^{H \times W}$, which avoids the hard boundaries introduced by binary masking and provides continuous local variation for subsequent mixing.

\subsubsection{Stochastic Non-Linear Shaping}
To further diversify the spatial distribution of mixing weights, we apply a stochastic shaping function to the interpolated surface: $M=\mathrm{Clip}\!\left(S^{\gamma},0,1\right), \gamma \sim \mathcal{U}(0.5,2.0)$, where $M \in [0,1]^{H \times W}$ is the final mixing mask. The exponent $\gamma$ controls the shape of the mask. When $\gamma<1$, the transformation increases values in $(0,1)$ and produces a smoother mask biased toward larger mixing weights; when $\gamma>1$, it suppresses smaller values and yields higher local contrast. We choose $\gamma\sim\mathcal{U}(0.5,2.0)$ because this range covers both smoothing and contrast-enhancing regimes while avoiding extreme exponents that make the mask nearly uniform or overly sparse. Empirically, fixing $\gamma=1$ decreases the CIFAR-100-LT accuracy from 41.04\% to 40.35\%, indicating that stochastic shaping contributes beyond bicubic smoothing alone.

\subsubsection{Element-wise Mixing and Label Reweighting}
Given the spatial mask $M$, TopoMix performs element-wise mixing: $\tilde{\mathbf{x}} = M \odot \mathbf{x}_A + (1-M) \odot \mathbf{x}_B$, where $\odot$ denotes the Hadamard product. The effective mixing ratio is defined as the spatial average of the mask: $\lambda_{\text{eff}}=\frac{1}{H \times W}\sum_{u,v} M_{u,v}$, and the mixed label is computed as $\tilde{y}=\lambda_{\text{eff}} y_A + (1-\lambda_{\text{eff}}) y_B$.

\begin{figure*}[t]
	\centering
	\vspace*{5pt}
	\includegraphics[width=0.99\textwidth]{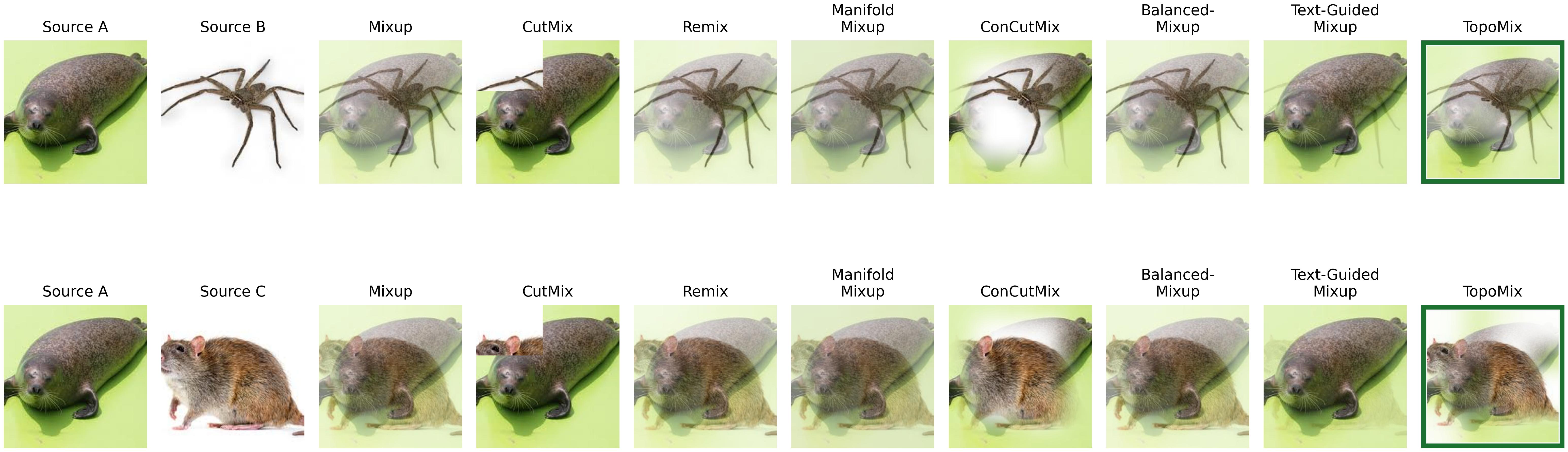}
	\caption{Qualitative comparison of mixing results. Compared to the ghosting of Mixup and the sharp boundaries of CutMix, TopoMix achieves smooth spatial transitions and more coherent structure--texture blending.}
	\label{fig:Vis}
\end{figure*}

Unlike global linear interpolation, this formulation allows different spatial locations to take different mixing weights while keeping the overall process simple and training-free. Compared with Mixup and CutMix, TopoMix provides smoother spatial transitions and more flexible local blending, as illustrated in Fig.~\ref{fig:Vis}. Algorithm~\ref{alg:topomix} summarizes the full TopoMix augmentation procedure, including control-grid sampling, bicubic upsampling, stochastic shaping, element-wise mixing, and label reweighting. This algorithmic view complements Fig.~\ref{fig:frame} by making the sample generation process explicit step by step.

\begin{table}[t]
	\vspace*{5pt}
	\caption{Top-1 accuracy (\%) on CIFAR-10-LT, CIFAR-100-LT, and ImageNet-LT.}
	\label{tab:summary}
	\centering
	\setlength{\tabcolsep}{4pt}
	\footnotesize
	\resizebox{\columnwidth}{!}{
		\begin{tabular}{l c c c}
			\toprule
			\multirow{2}{*}{\textbf{Method}} & \textbf{CIFAR-10-LT} & \textbf{CIFAR-100-LT} & \textbf{ImageNet-LT} \\
			& (IF=100) & (IF=100) & (ResNet-32) \\
			\midrule
			Baseline (ResNet-32) & 69.65 & 39.19 & 38.30 \\
			Mixup \cite{zhang2017mixup} & 70.72 & 39.45 & 39.30 \\
			CutMix \cite{yun2019cutmix} & 69.89 & 39.23 & 38.50 \\
			Remix \cite{chou2020remix} & 71.32 & 40.24 & 39.71 \\
			Balanced-Mixup \cite{galdran2021balanced} & 71.85 & 40.68 & 40.25 \\
			Manifold Mixup \cite{verma2019manifold} & 70.95 & 39.82 & 39.55 \\
			Text-Guided Mixup \cite{DBLP:conf/bmvc/FranklinYZ0H24} & 72.15 & 40.85 & 40.80 \\
			ConCutMix \cite{10599153} & 71.60 & 40.55 & 40.10 \\
			\midrule
			\rowcolor{gray!12}
			\textbf{TopoMix (Ours)} & \textbf{72.47} & \textbf{41.04} & \textbf{41.14} \\
			\bottomrule
		\end{tabular}
	}
\end{table}

\begin{table}[t]
	\caption{Detailed comparison on ImageNet-LT under ResNet-32 and ResNet-50 backbones.}
	\label{tab:imagenet_detail}
	\centering
	\setlength{\tabcolsep}{4pt}
	\footnotesize
	\resizebox{\columnwidth}{!}{
		\begin{tabular}{l c c c c c}
			\toprule
			\textbf{Method} & \textbf{Backbone} & \textbf{Total} & \textbf{Head} & \textbf{Med} & \textbf{Tail} \\
			\midrule
			\multicolumn{6}{c}{\textit{ResNet-32 backbone}} \\
			\midrule
			Baseline & ResNet-32 & 38.30 & 61.73 & 30.23 & 4.96 \\
			Mixup \cite{zhang2017mixup} & ResNet-32 & 39.30 & 62.88 & 31.95 & 5.14 \\
			CutMix \cite{yun2019cutmix} & ResNet-32 & 38.50 & 62.15 & 30.75 & 4.98 \\
			Remix \cite{chou2020remix} & ResNet-32 & 39.71 & 62.36 & 32.52 & 5.85 \\
			Manifold Mixup \cite{verma2019manifold} & ResNet-32 & 39.55 & 62.55 & 32.20 & 5.25 \\
			ConCutMix \cite{10599153} & ResNet-32 & 40.10 & 62.60 & 32.85 & 5.40 \\
			Balanced-Mixup \cite{galdran2021balanced} & ResNet-32 & 40.25 & 61.90 & 33.15 & 5.70 \\
			Text-Guided Mixup \cite{DBLP:conf/bmvc/FranklinYZ0H24} & ResNet-32 & 40.80 & 62.90 & 33.35 & 5.95 \\
			\rowcolor{gray!12}
			\textbf{TopoMix (Ours)} & \textbf{ResNet-32} & \textbf{41.14} & \textbf{63.95} & \textbf{33.60} & \textbf{6.10} \\
			\midrule
			\multicolumn{6}{c}{\textit{ResNet-50 backbone}} \\
			\midrule
			Baseline & ResNet-50 & 40.60 & 63.82 & 33.42 & 5.56 \\
			ConCutMix \cite{10599153} & ResNet-50 & 42.50 & 64.90 & 35.20 & 6.45 \\
			Text-Guided Mixup \cite{DBLP:conf/bmvc/FranklinYZ0H24} & ResNet-50 & 43.15 & 65.25 & 35.80 & 6.85 \\
			\rowcolor{gray!12}
			\textbf{TopoMix (Ours)} & \textbf{ResNet-50} & \textbf{44.20} & \textbf{66.30} & \textbf{36.55} & \textbf{7.50} \\
			\bottomrule
		\end{tabular}
	}
\end{table}

\section{Experiments}

\subsection{Experimental Setup}

Datasets.
We evaluate TopoMix on three standard long-tailed benchmarks covering both small-scale and large-scale settings.
\begin{itemize}
	\item CIFAR-10-LT and CIFAR-100-LT: Long-tailed versions of CIFAR constructed with imbalance factor (IF) 100.
	\item ImageNet-LT: A large-scale long-tailed benchmark containing 115.8K images from 1,000 categories, with class frequencies ranging from 5 to 1,280.
\end{itemize}

Backbone and hardware. Unless otherwise stated, we use ResNet-32 \cite{he2016deep} for the main comparison. For CIFAR-10-LT and CIFAR-100-LT, we adopt the standard ResNet-32 designed for $32\times32$ inputs. For ImageNet-LT, we use a ResNet-32 variant adapted to $224\times224$ inputs. To further examine whether the proposed mixing strategy generalizes beyond the main backbone, we additionally report ResNet-50 results on ImageNet-LT. CIFAR experiments are conducted on a local workstation with one NVIDIA RTX 4070 (8GB). ImageNet-LT experiments are conducted on a cloud server with one NVIDIA A100 (80GB) GPU.

Training protocol. Unless otherwise stated, all models are trained for 100 epochs using SGD with momentum 0.9 and weight decay $5\times10^{-4}$. We use cosine annealing with an initial learning rate of 0.1. For TopoMix, the control-grid size is set to $k=4$, the non-linear exponent is sampled from $\gamma\sim\mathcal{U}(0.5,2.0)$, and the mixing probability is set to $p=1.0$ by default. Baselines and implementation notes. We compare TopoMix with a broad set of representative mixing-based baselines, including Mixup \cite{zhang2017mixup}, CutMix \cite{yun2019cutmix}, Remix \cite{chou2020remix}, Balanced-Mixup \cite{galdran2021balanced}, Manifold Mixup \cite{verma2019manifold}, Text-Guided Mixup \cite{DBLP:conf/bmvc/FranklinYZ0H24}, and ConCutMix \cite{10599153}. For fair comparison, all methods are re-run under the same data split, backbone, and training schedule on each benchmark; all mixing methods use the same randomly paired training images, and the mixed samples are merged with the original images and shuffled for training. Balanced-Mixup is adapted from its original imbalanced classification setting by replacing the standard mixing pair construction with class-balanced sampling. Manifold Mixup performs interpolation in hidden space rather than the input space. Text-Guided Mixup and ConCutMix are also evaluated under the same protocol, since their original implementations may use benchmark-specific settings or different backbone choices.

\subsection{Main Results: Consistency Across Benchmarks}
Table~\ref{tab:summary} reports the main comparison on CIFAR-10-LT, CIFAR-100-LT, and ImageNet-LT. TopoMix achieves the best performance on all three benchmarks. Compared with the second-best method, namely Text-Guided Mixup, TopoMix improves accuracy by 0.32\% on CIFAR-10-LT, 0.19\% on CIFAR-100-LT, and 0.34\% on ImageNet-LT. Relative to the plain ResNet-32 baseline, the gains reach 2.82\%, 1.85\%, and 2.84\%, respectively. These results show that replacing fixed global mixing with a smooth spatially varying mask remains effective across both small-scale and large-scale long-tailed settings.

\begin{table}[t]
	\caption{Ablation and hyperparameter sensitivity on ImageNet-LT with ResNet-50. Accuracy (\%). Default: $k=4$, $\gamma\sim\mathcal{U}(0.5,2.0)$, $p=1.0$.}
	\label{tab:ablation_imagenet_r50}
	\centering
	\setlength{\tabcolsep}{5pt}
	\footnotesize
	\resizebox{\columnwidth}{!}{
		\begin{tabular}{l c c c c}
			\toprule
			\textbf{Setting} & \textbf{Total} & \textbf{Head} & \textbf{Med} & \textbf{Tail} \\
			\midrule
			Baseline (ResNet-50) & 40.60 & 63.82 & 33.42 & 5.56 \\
			TopoMix (default) $(k=4,\, p=1.0)$ & 44.20 & 66.30 & 36.55 & 7.50 \\
			\midrule
			w/o shaping ($\gamma=1$, $k=4$, $p=1.0$) & 43.05 & 66.00 & 35.15 & 6.20 \\
			\midrule
			$k=2$ (with $p=1.0$) & 42.75 & 66.15 & 34.85 & 5.95 \\
			$k=4$ (with $p=1.0$) & 44.20 & 66.30 & 36.55 & 7.50 \\
			$k=8$ (with $p=1.0$) & 43.45 & 65.40 & 35.80 & 6.90 \\
			\midrule
			$p=0.5$ (with $k=4$) & 43.65 & 66.45 & 35.90 & 6.60 \\
			$p=1.0$ (with $k=4$) & 44.20 & 66.30 & 36.55 & 7.50 \\
			\bottomrule
		\end{tabular}
	}
\end{table}

\begin{figure}[t]
	\centering
	\vspace*{5pt}
	\includegraphics[width=0.99\linewidth]{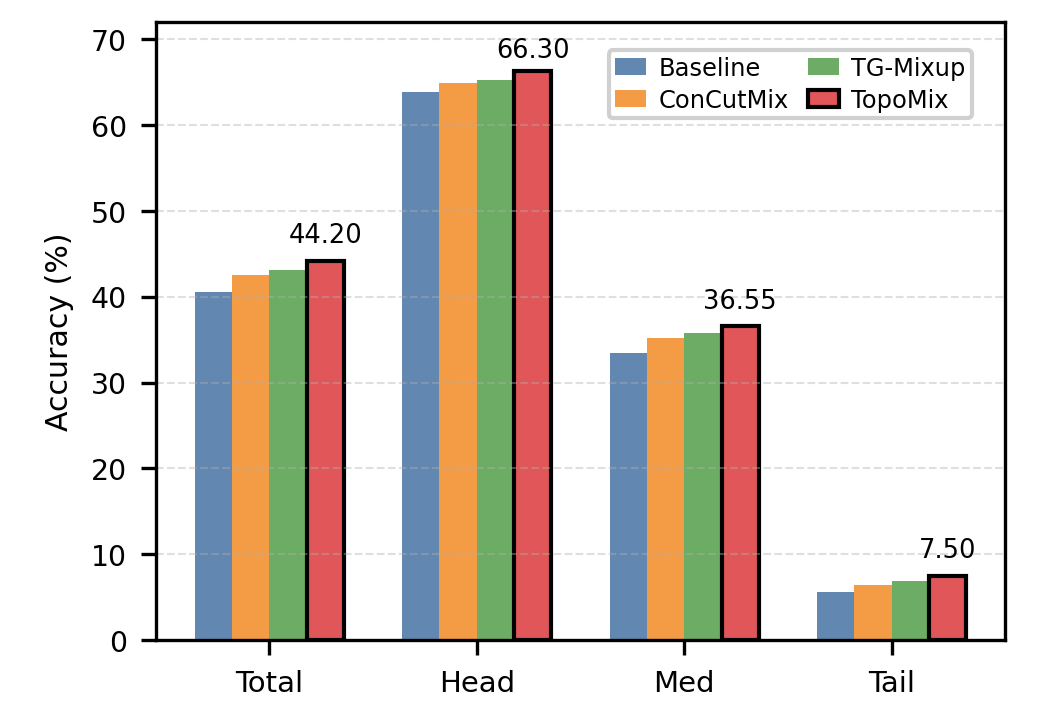}
	\caption{ImageNet-LT comparison under ResNet-50.}
	\label{fig:imagenet_r50_bar}
\end{figure}

\subsection{Ablation Studies and Hyperparameter Sensitivity}
We conduct ablation studies to examine the key design choices in TopoMix.

Ablation on CIFAR-100-LT.
Table~\ref{tab:ablation} presents the ablation results on CIFAR-100-LT with ResNet-32. Unless otherwise stated, the default setting is $k=4$, $\gamma\sim\mathcal{U}(0.5,2.0)$, and $p=1.0$.
Removing the shaping stage by fixing $\gamma=1$ reduces accuracy from 41.04\% to 40.35\%, showing that stochastic non-linear shaping contributes beyond smooth upsampling alone. Varying the control-grid size further shows that $k=4$ gives the best result, while $k=2$ is overly coarse and $k=8$ introduces stronger local variation that is less effective. We also compare $p=0.5$ and $p=1.0$, and find that full application of TopoMix yields the strongest performance. Overall, the results indicate that TopoMix is not overly sensitive to its hyperparameters and that the default configuration provides a stable operating point.

Ablation on ImageNet-LT with ResNet-50.
To verify whether the same trends remain valid on a stronger backbone and a larger benchmark, we further report ablation results on ImageNet-LT with ResNet-50 in Table~\ref{tab:ablation_imagenet_r50}. The observations are consistent with those on CIFAR-100-LT. Removing stochastic shaping degrades performance across Total, Med, and Tail accuracy. The default setting $k=4$ again performs best, while both smaller and larger control grids are less effective. In addition, using $p=1.0$ remains better than $p=0.5$. These results suggest that the main design choices of TopoMix transfer consistently from CIFAR-LT to ImageNet-LT and from ResNet-32 to ResNet-50.

\subsection{Cross-Architecture Comparison on ImageNet-LT}

Table~\ref{tab:imagenet_detail} provides a more detailed comparison on ImageNet-LT under both ResNet-32 and ResNet-50 backbones. Under ResNet-32, TopoMix achieves the best overall performance and improves the tail accuracy from 4.96\% to 6.10\% over the baseline. Under ResNet-50, the gains remain clear: compared with the ResNet-50 baseline, TopoMix improves Total accuracy from 40.60\% to 44.20\%, and improves Tail accuracy from 5.56\% to 7.50\%. It also outperforms the strongest competing method, Text-Guided Mixup, by 1.05\% in Total accuracy and 0.65\% in Tail accuracy. Fig.~\ref{fig:imagenet_r50_bar} gives a clearer visual comparison of the ResNet-50 results. These results indicate that the proposed mixing strategy is effective not only on the main ResNet-32 setting but also on a stronger and more widely used backbone. Importantly, TopoMix does not trade head-class accuracy for tail-class gains, suggesting that its advantage comes from improved sample mixing quality rather than a bias toward minority classes alone.

\subsection{Feature Visualization}
\begin{figure}[t]
	\centering
	\vspace*{5pt}
	\includegraphics[width=0.99\linewidth]{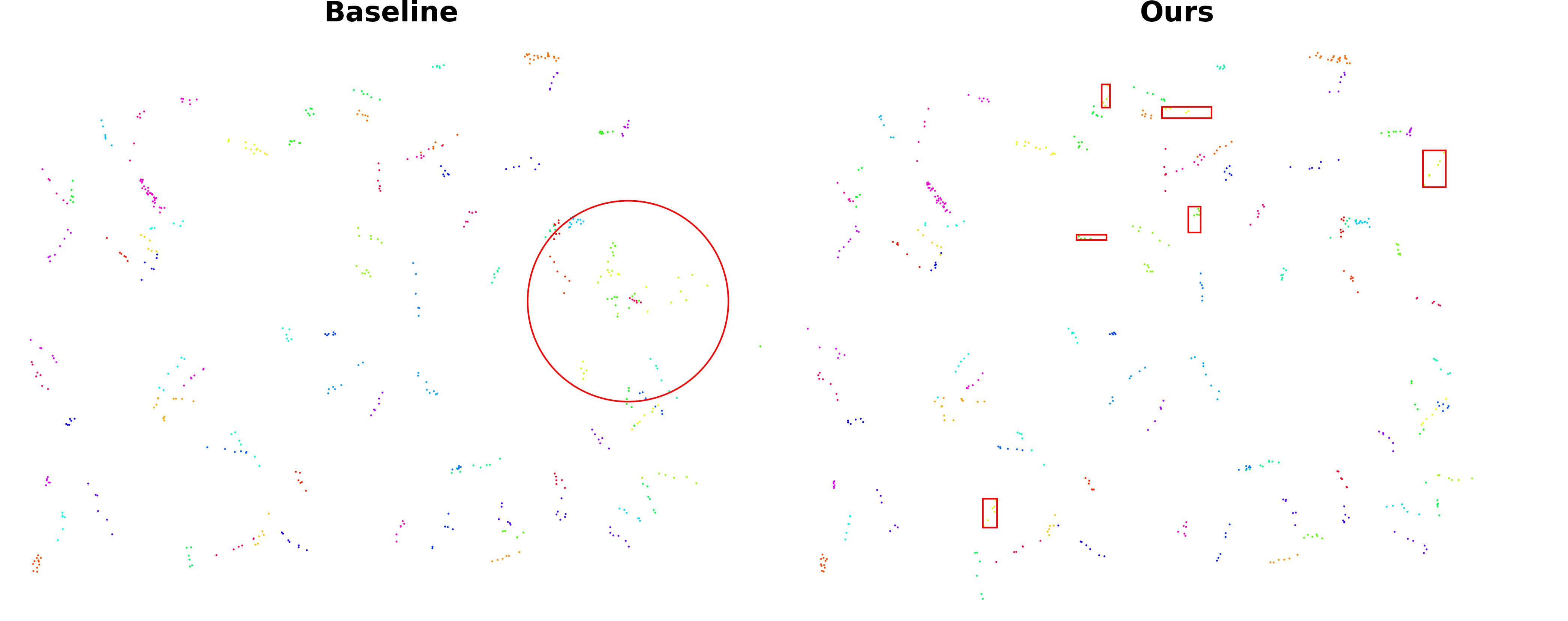}
	\caption{t-SNE visualization on CIFAR-100-LT. Compared with the baseline(ResNet-32), TopoMix yields more compact features and clearer tail-class clusters.}
	\label{fig:compare}
\end{figure}

Fig.~\ref{fig:compare} provides a qualitative view of the learned features on CIFAR-100-LT. Compared with the baseline, TopoMix produces more compact feature clusters and clearer tail-class separation, which is consistent with its quantitative improvements on minority classes.

\begin{table}[t]
	\caption{Ablation and hyperparameter sensitivity on CIFAR-100-LT (IF=100) with ResNet-32. Top-1 accuracy (\%). Default: $k=4$, $\gamma\sim\mathcal{U}(0.5,2.0)$, $p=1.0$.}
	\label{tab:ablation}
	\centering
	\setlength{\tabcolsep}{5pt}
	\begin{tabular}{l c}
		\toprule
		\textbf{Setting} & \textbf{Top-1 (\%)} \\
		\midrule
		Baseline (ResNet-32) & 39.19 \\
		TopoMix (default) $(k=4,\, p=1.0)$ & 41.04 \\
		\midrule
		w/o shaping ($\gamma=1$, $k=4$, $p=1.0$) & 40.35 \\
		\midrule
		$k=2$ (with $p=1.0$) & 40.20 \\
		$k=4$ (with $p=1.0$) & 41.04 \\
		$k=8$ (with $p=1.0$) & 40.60 \\
		\midrule
		$p=0.5$ (with $k=4$) & 40.70 \\
		$p=1.0$ (with $k=4$) & 41.04 \\
		\bottomrule
	\end{tabular}
\end{table}

\section{Conclusion}
\label{sec:conclusion}

In this paper, we presented TopoMix, a training-free data augmentation method for long-tailed recognition based on continuous spatial mixing. Instead of relying on a single global interpolation ratio or hard regional replacement, TopoMix generates a smooth spatially varying mask from a random control grid, bicubic interpolation, and stochastic non-linear shaping, and then performs element-wise mixing with label reweighting. This design makes the mixing process more flexible in the input space while preserving a simple overall pipeline. Experiments on standard long-tailed benchmarks show that TopoMix consistently improves over representative mixing-based baselines. The gains are observed not only in overall accuracy, but also more clearly on minority classes, suggesting that smooth spatial mixing is a useful alternative to fixed global mixing in long-tailed settings. Further comparisons on both ResNet-32 and ResNet-50 indicate that the proposed strategy remains effective across different backbone choices. In addition, the ablation results consistently support the roles of stochastic shaping, moderate control-grid size, and full mixing probability in the final performance. Overall, these results suggest that improving how samples are mixed can be as important as adjusting how samples are reweighted. By replacing fixed-ratio mixing with continuous local variation, TopoMix provides a simple and effective way to improve long-tailed recognition.

\bibliographystyle{IEEEtran}
\bibliography{rootreferences}

@article{zhang2017mixup,
	title={mixup: Beyond Empirical Risk Minimization},
    author={Hongyi Zhang and Moustapha Cisse and Yann N. Dauphin and David Lopez-Paz},
    booktitle={International Conference on Learning Representations},
    year={2018},
    url={https://openreview.net/forum?id=r1Ddp1-Rb},
}

@inproceedings{yun2019cutmix,
author={Yun, Sangdoo and Han, Dongyoon and Chun, Sanghyuk and Oh, Seong Joon and Yoo, Youngjoon and Choe, Junsuk},
booktitle={2019 IEEE/CVF International Conference on Computer Vision (ICCV)}, 
title={CutMix: Regularization Strategy to Train Strong Classifiers With Localizable Features}, 
year={2019},
volume={},
number={},
pages={6022-6031},
doi={10.1109/ICCV.2019.00612}}

@InProceedings{verma2019manifold,
	title = 	 {Manifold Mixup: Better Representations by Interpolating Hidden States},
	author =       {Verma, Vikas and Lamb, Alex and Beckham, Christopher and Najafi, Amir and Mitliagkas, Ioannis and Lopez-Paz, David and Bengio, Yoshua},
	booktitle = 	 {Proceedings of the 36th International Conference on Machine Learning},
	pages = 	 {6438--6447},
	year = 	 {2019},
	editor = 	 {Chaudhuri, Kamalika and Salakhutdinov, Ruslan},
	volume = 	 {97},
	series = 	 {Proceedings of Machine Learning Research},
	month = 	 {09--15 Jun},
	publisher =    {PMLR},
	url = 	 {https://proceedings.mlr.press/v97/verma19a.html}
}

@InProceedings{cubuk2019autoaugment,
	author = {Cubuk, Ekin D. and Zoph, Barret and Mane, Dandelion and Vasudevan, Vijay and Le, Quoc V.},
	title = {AutoAugment: Learning Augmentation Strategies From Data},
	booktitle = {Proceedings of the IEEE/CVF Conference on Computer Vision and Pattern Recognition (CVPR)},
	month = {June},
	year = {2019}
}

@misc{fefferman2016testing,
	title={Testing the Manifold Hypothesis}, 
	author={Charles Fefferman and Sanjoy Mitter and Hariharan Narayanan},
	year={2013},
	eprint={1310.0425},
	archivePrefix={arXiv},
	primaryClass={math.ST},
	url={https://arxiv.org/abs/1310.0425}, 
}

@inproceedings{narayanan2010sample,
	author = {Narayanan, Hariharan and Mitter, Sanjoy},
	booktitle = {Advances in Neural Information Processing Systems},
	editor = {J. Lafferty and C. Williams and J. Shawe-Taylor and R. Zemel and A. Culotta},
	pages = {},
	publisher = {Curran Associates, Inc.},
	title = {Sample Complexity of Testing the Manifold Hypothesis},
	url = {https://proceedings.neurips.cc/paper_files/paper/2010/file/8a1e808b55fde9455cb3d8857ed88389-Paper.pdf},
	volume = {23},
	year = {2010}
}

@article{fukunaga1971algorithm,
author={Fukunaga, K. and Olsen, D.R.},
journal={IEEE Transactions on Computers}, 
title={An Algorithm for Finding Intrinsic Dimensionality of Data}, 
year={1971},
volume={C-20},
number={2},
pages={176-183},
doi={10.1109/T-C.1971.223208}}

@InProceedings{gong2019intrinsic,
	author = {Gong, Sixue and Boddeti, Vishnu Naresh and Jain, Anil K.},
	title = {On the Intrinsic Dimensionality of Image Representations},
	booktitle = {Proceedings of the IEEE/CVF Conference on Computer Vision and Pattern Recognition (CVPR)},
	month = {June},
	year = {2019}
}

@inproceedings{pope2021intrinsic,
	title={The Intrinsic Dimension of Images and Its Impact on Learning},
	author={Phil Pope and Chen Zhu and Ahmed Abdelkader and Micah Goldblum and Tom Goldstein},
	booktitle={International Conference on Learning Representations},
	year={2021},
	url={https://openreview.net/forum?id=XJk19XzGq2J}
}

@inproceedings{
	kang2019decoupling,
	title={Decoupling Representation and Classifier for Long-Tailed Recognition},
	author={Bingyi Kang and Saining Xie and Marcus Rohrbach and Zhicheng Yan and Albert Gordo and Jiashi Feng and Yannis Kalantidis},
	booktitle={International Conference on Learning Representations},
	year={2020},
	url={https://openreview.net/forum?id=r1gRTCVFvB}
}

@InProceedings{liu2019large,
	author = {Liu, Ziwei and Miao, Zhongqi and Zhan, Xiaohang and Wang, Jiayun and Gong, Boqing and Yu, Stella X.},
	title = {Large-Scale Long-Tailed Recognition in an Open World},
	booktitle = {Proceedings of the IEEE/CVF Conference on Computer Vision and Pattern Recognition (CVPR)},
	month = {June},
	year = {2019}
}

@inproceedings{alshammari2022long,
author={Alshammari, Shaden and Wang, Yu-Xiong and Ramanan, Deva and Kong, Shu},
booktitle={2022 IEEE/CVF Conference on Computer Vision and Pattern Recognition (CVPR)}, 
title={Long- Tailed Recognition via Weight Balancing}, 
year={2022},
volume={},
number={},
pages={6887-6897},
doi={10.1109/CVPR52688.2022.00677}}

@inproceedings{cubuk2020randaugment,
author = {Cubuk, Ekin Dogus and Zoph, Barret and Shlens, Jon and Le, Quoc},
booktitle = {Advances in Neural Information Processing Systems},
editor = {H. Larochelle and M. Ranzato and R. Hadsell and M.F. Balcan and H. Lin},
pages = {18613--18624},
publisher = {Curran Associates, Inc.},
title = {RandAugment: Practical Automated Data Augmentation with a Reduced Search Space},
url = {https://proceedings.neurips.cc/paper_files/paper/2020/file/d85b63ef0ccb114d0a3bb7b7d808028f-Paper.pdf},
volume = {33},
year = {2020}
}

@inproceedings{chou2020remix,
	author="Chou, Hsin-Ping
	and Chang, Shih-Chieh
	and Pan, Jia-Yu
	and Wei, Wei
	and Juan, Da-Cheng",
	editor="Bartoli, Adrien
	and Fusiello, Andrea",
	title="Remix: Rebalanced Mixup",
	booktitle="Computer Vision -- ECCV 2020 Workshops",
	year="2020",
	publisher="Springer International Publishing",
	address="Cham",
	pages="95--110",
	isbn="978-3-030-65414-6"
	}

@article{cao2019learning,
author = {Cao, Kaidi and Wei, Colin and Gaidon, Adrien and Arechiga, Nikos and Ma, Tengyu},
booktitle = {Advances in Neural Information Processing Systems},
editor = {H. Wallach and H. Larochelle and A. Beygelzimer and F. d\textquotesingle Alch\'{e}-Buc and E. Fox and R. Garnett},
pages = {},
publisher = {Curran Associates, Inc.},
title = {Learning Imbalanced Datasets with Label-Distribution-Aware Margin Loss},
volume = {32},
year = {2019}
}

@inproceedings{he2016deep,
author={He, Kaiming and Zhang, Xiangyu and Ren, Shaoqing and Sun, Jian},
booktitle={2016 IEEE Conference on Computer Vision and Pattern Recognition (CVPR)}, 
title={Deep Residual Learning for Image Recognition}, 
year={2016},
volume={},
number={},
pages={770-778},
doi={10.1109/CVPR.2016.90}}

@inproceedings{dosovitskiy2020image,
	title={An Image is Worth 16x16 Words: Transformers for Image Recognition at Scale},
	author={Alexey Dosovitskiy and Lucas Beyer and Alexander Kolesnikov and Dirk Weissenborn and Xiaohua Zhai and Thomas Unterthiner and Mostafa Dehghani and Matthias Minderer and Georg Heigold and Sylvain Gelly and Jakob Uszkoreit and Neil Houlsby},
	booktitle={International Conference on Learning Representations},
	year={2021},
	url={https://openreview.net/forum?id=YicbFdNTTy}
}

@inproceedings{krizhevsky2009learning,
	title={Learning Multiple Layers of Features from Tiny Images},
	author={Alex Krizhevsky},
	year={2009},
	url={https://api.semanticscholar.org/CorpusID:18268744}
}

@inproceedings{deng2009imagenet,
author={Deng, Jia and Dong, Wei and Socher, Richard and Li, Li-Jia and Kai Li and Li Fei-Fei},
booktitle={2009 IEEE Conference on Computer Vision and Pattern Recognition}, 
title={ImageNet: A large-scale hierarchical image database}, 
year={2009},
volume={},
number={},
pages={248-255},
doi={10.1109/CVPR.2009.5206848}}

@inproceedings{galdran2021balanced,
author="Galdran, Adrian
and Carneiro, Gustavo
and Gonz{\'a}lez Ballester, Miguel A.",
editor="de Bruijne, Marleen
and Cattin, Philippe C.
and Cotin, St{\'e}phane
and Padoy, Nicolas
and Speidel, Stefanie
and Zheng, Yefeng
and Essert, Caroline",
title="Balanced-MixUp for Highly Imbalanced Medical Image Classification",
booktitle="Medical Image Computing and Computer Assisted Intervention -- MICCAI 2021",
year="2021",
publisher="Springer International Publishing",
address="Cham",
pages="323--333",
isbn="978-3-030-87240-3"
}

@article{ni2024survey,
title={A Survey on Multimodal Wearable Sensor-based Human Action Recognition}, 
author={Jianyuan Ni and Hao Tang and Syed Tousiful Haque and Yan Yan and Anne H. H. Ngu},
year={2024},
eprint={2404.15349},
archivePrefix={arXiv},
primaryClass={eess.SP},
url={https://arxiv.org/abs/2404.15349}, 
}

@article{maray2023transfer,
AUTHOR = {Maray, Nader and Ngu, Anne Hee and Ni, Jianyuan and Debnath, Minakshi and Wang, Lu},
TITLE = {Transfer Learning on Small Datasets for Improved Fall Detection},
JOURNAL = {Sensors},
VOLUME = {23},
YEAR = {2023},
NUMBER = {3},
ARTICLE-NUMBER = {1105},
URL = {https://www.mdpi.com/1424-8220/23/3/1105},
PubMedID = {36772148},
ISSN = {1424-8220},
DOI = {10.3390/s23031105}
}

@inproceedings{ni2025s3,
author={Ni, Jianyuan and Sarbajna, Raunak and Ngu, Anne H.H.},
booktitle={2025 IEEE International Conference on Pervasive Computing and Communications Workshops and other Affiliated Events (PerCom Workshops)}, 
title={S3 KD: Semantic-aware Spiking Neural Networks Self-distillation for Human Action Recognition}, 
year={2025},
volume={},
number={},
pages={92-97},
doi={10.1109/PerComWorkshops65533.2025.00047}}

@article{ding2025advances,
title = {Advances in deep learning for personalized ECG diagnostics: A systematic review addressing inter-patient variability and generalization constraints},
journal = {Biosensors and Bioelectronics},
volume = {271},
pages = {117073},
year = {2025},
issn = {0956-5663},
doi = {https://doi.org/10.1016/j.bios.2024.117073},
url = {https://www.sciencedirect.com/science/article/pii/S0956566324010807},
author = {Cheng Ding and Tianliang Yao and Chenwei Wu and Jianyuan Ni}
}

@article{qin2025kedformer,
	title={KEDformer: Knowledge extraction seasonal trend decomposition for long-term sequence prediction},
	author={Qin, Zhenkai and Wei, Baozhong and Gao, Caifeng and Ni, Jianyuan},
	journal={PLoS One},
	volume={20},
	number={10},
	pages={e0335047},
	year={2025},
	publisher={Public Library of Science San Francisco, CA USA}
}

@article{feng2025machine,
	title={Machine learning to assess and support safe drinking water supply: a systematic review},
	author={Feng, Feng and Zhang, Yuanxun and Chen, Zhenru and Ni, Jianyuan and Feng, Yuan and Xie, Yunchao and Zhang, Chiqian},
	journal={Journal of Environmental Chemical Engineering},
	volume={13},
	number={1},
	pages={114481},
	year={2025},
	publisher={Elsevier}
}

@inproceedings{ni2022cross,
	title={Cross-modal knowledge distillation for vision-to-sensor action recognition},
	author={Ni, Jianyuan and Sarbajna, Raunak and Liu, Yang and Ngu, Anne HH and Yan, Yan},
	booktitle={ICASSP},
	pages={4448--4452},
	year={2022},
	organization={IEEE}
}

@inproceedings{ni2022progressive,
	title={Progressive cross-modal knowledge distillation for human action recognition},
	author={Ni, Jianyuan and Ngu, Anne HH and Yan, Yan},
	booktitle={ACM MM},
	pages={5903--5912},
	year={2022}
}

@INPROCEEDINGS{8953804,
	author = {Cui, Yin and Jia, Menglin and Lin, Tsung-Yi and Song, Yang and Belongie, Serge},
	title = {Class-Balanced Loss Based on Effective Number of Samples},
	booktitle = {Proceedings of the IEEE/CVF Conference on Computer Vision and Pattern Recognition (CVPR)},
	month = {June},
	year = {2019}
}

@article{ren2020balanced,
author = {Ren, Jiawei and Yu, Cunjun and sheng, shunan and Ma, Xiao and Zhao, Haiyu and Yi, Shuai and Li, hongsheng},
booktitle = {Advances in Neural Information Processing Systems},
editor = {H. Larochelle and M. Ranzato and R. Hadsell and M.F. Balcan and H. Lin},
pages = {4175--4186},
publisher = {Curran Associates, Inc.},
title = {Balanced Meta-Softmax for Long-Tailed Visual Recognition},
url = {https://proceedings.neurips.cc/paper_files/paper/2020/file/2ba61cc3a8f44143e1f2f13b2b729ab3-Paper.pdf},
volume = {33},
year = {2020}
}

@inproceedings{DBLP:conf/bmvc/FranklinYZ0H24,
author    = {Richard Franklin and Jiawei Yao and Deyang Zhong and Qi Qian and Juhua Hu},
title     = {Text-Guided Mixup Towards Long-Tailed Image Categorization},
booktitle = {35th British Machine Vision Conference 2024, {BMVC} 2024, Glasgow, UK, November 25-28, 2024},
publisher = {BMVA},
year      = {2024}
}

@ARTICLE{10599153,
author={Pan, Haolin and Guo, Yong and Yu, Mianjie and Chen, Jian},
journal={IEEE Transactions on Image Processing}, 
title={Enhanced Long-Tailed Recognition With Contrastive CutMix Augmentation}, 
year={2024},
volume={33},
number={},
pages={4215-4230},
doi={10.1109/TIP.2024.3425148}}

@InProceedings{Zhou_2020_CVPR,
author = {Zhou, Boyan and Cui, Quan and Wei, Xiu-Shen and Chen, Zhao-Min},
title = {BBN: Bilateral-Branch Network With Cumulative Learning for Long-Tailed Visual Recognition},
booktitle = {Proceedings of the IEEE/CVF Conference on Computer Vision and Pattern Recognition (CVPR)},
month = {June},
year = {2020}
}

@InProceedings{Hong_2021_CVPR,
	author    = {Hong, Youngkyu and Han, Seungju and Choi, Kwanghee and Seo, Seokjun and Kim, Beomsu and Chang, Buru},
	title     = {Disentangling Label Distribution for Long-Tailed Visual Recognition},
	booktitle = {Proceedings of the IEEE/CVF Conference on Computer Vision and Pattern Recognition (CVPR)},
	month     = {June},
	year      = {2021},
	pages     = {6626-6636}
}

@InProceedings{Aimar_2023_CVPR,
author={Aimar, Emanuel Sanchez and Jonnarth, Arvi and Felsberg, Michael and Kuhlmann, Marco},
booktitle={2023 IEEE/CVF Conference on Computer Vision and Pattern Recognition (CVPR)}, 
title={Balanced Product of Calibrated Experts for Long-Tailed Recognition}, 
year={2023},
volume={},
number={},
pages={19967-19977},
doi={10.1109/CVPR52729.2023.01912}}

@inproceedings{DBLP:conf/iclr/MenonJRJVK21,
	title={Long-tail learning via logit adjustment},
	author={Aditya Krishna Menon and Sadeep Jayasumana and Ankit Singh Rawat and Himanshu Jain and Andreas Veit and Sanjiv Kumar},
	booktitle={International Conference on Learning Representations},
	year={2021}
}

@inproceedings{DBLP:conf/iclr/WangLM0Y21,
	title={Long-tailed Recognition by Routing Diverse Distribution-Aware Experts},
	author={Xudong Wang and Long Lian and Zhongqi Miao and Ziwei Liu and Stella Yu},
	booktitle={International Conference on Learning Representations},
	year={2021}
}

\end{document}